\documentclass[conference]{IEEEtran}
\IEEEoverridecommandlockouts
\usepackage{cite}
\usepackage{amsmath,amsfonts}
\usepackage{graphicx}
\usepackage{textcomp}
\usepackage{xcolor}
\usepackage{balance}
\usepackage[ruled, linesnumbered, commentsnumbered, longend]{algorithm2e}
\usepackage{algpseudocode}
\usepackage{array}
\usepackage{subcaption}
\usepackage{float} 
\usepackage{makecell}
\usepackage{booktabs}
\usepackage{url}
\usepackage{flushend}
\usepackage{tikz,tkz-euclide,pdftexcmds,calc}
\usepackage{placeins}
\usetikzlibrary{arrows,shapes.geometric}

\usepackage{amsmath}
\usepackage{amsfonts}
\usepackage{mathtools}
\usepackage{bm}

\def\BibTeX{{\rm B\kern-.05em{\sc i\kern-.025em b}\kern-.08em
    T\kern-.1667em\lower.7ex\hbox{E}\kern-.125emX}}
\begin{document}

\title{Advancing Intoxication Detection: A Smartwatch-Based Approach}

\newcommand{\linebreakand}{%
  \end{@IEEEauthorhalign}
  \hfill\mbox{}\par
  \mbox{}\hfill\begin{@IEEEauthorhalign}
}

\author{\IEEEauthorblockN{Manuel E. Segura\textsuperscript{\textsection}\textsuperscript{*},
 Pere Vergés\textsuperscript{\textsection}\textsuperscript{*}, Richard Ky\textsuperscript{*}, Ramesh Arangott\textsuperscript{+}, Angela Kristine Garcia\textsuperscript{+}, \\ 
 Thang Dinh Trong\textsuperscript{+}, Makoto Hyodo\textsuperscript{+}, Alexandru Nicolau\textsuperscript{*}, Tony Givargis\textsuperscript{*} and Sergio Gago-Masague\textsuperscript{*}
}
\IEEEauthorblockA{University of California Irvine\textsuperscript{*}, Asahi Group Holdings Ltd.\textsuperscript{+}\\
mesegur1@uci.edu, pvergesb@uci.edu, kyr@uci.edu, ramesh.arangott@asahigroup-holdings.com, \\
angela.kristinegarcia@asahigroup-holdings.com, thang.dinhtrong@asahigroup-holdings.com, \\
makoto.hyodo@asahigroup-holdings.com, nicolau@ics.uci.edu, givargis@uci.edu, sgagomas@uci.edu}}

\maketitle
\begingroup\renewcommand\thefootnote{\textsection}
\footnotetext{Both authors contributed equally to this research.}

\begin{abstract}
Excess alcohol consumption leads to serious health risks and severe consequences for both individuals and their communities. To advocate for healthier drinking habits, we introduce a groundbreaking mobile smartwatch application approach to just-in-time interventions for intoxication warnings. In this work, we have created a dataset gathering TAC, accelerometer, gyroscope, and heart rate data from the participants during a period of three weeks. This is the first study to combine accelerometer, gyroscope, and heart rate smartwatch data collected over an extended monitoring period to classify intoxication levels. Previous research had used limited smartphone motion data and conventional machine learning (ML) algorithms to classify heavy drinking episodes; in this work, we use smartwatch data and perform a thorough evaluation of different state-of-the-art classifiers such as the Transformer, Bidirectional Long Short-Term Memory (bi-LSTM), Gated Recurrent Unit (GRU), One-Dimensional Convolutional Neural Networks (1D-CNN), and Hyperdimensional Computing (HDC). We have compared performance metrics for the algorithms and assessed their efficiency on resource-constrained environments like mobile hardware. The HDC model achieved the best balance between accuracy and efficiency, demonstrating its practicality for smartwatch-based applications.
\end{abstract}


\section{Introduction}
The consumption of alcohol continues to pose a significant public health challenge, contributing to more than 200 diseases, injuries, and health conditions~\cite{alcohol}. In 2023, alcohol abuse accounted for 6\% of global deaths, with nearly 14\% of these fatalities occurring among individuals aged 20 to 39 years~\cite{drugabuse_alcohol_stats}. This alarming statistic underscores the profound impact of alcohol on human health, contributing to conditions such as liver cirrhosis, cancer, and pancreatitis~\cite{alchol_burden}. 
Additionally, behavioral changes induced by alcohol, such as those leading to motor vehicle accidents and episodes of anger~\cite{alcholmypoia}, further compound its detrimental effects.

Traditional methods for detecting intoxication, including blood, urine, and saliva tests, as well as devices such as breathalyzers, are intrusive, limiting their applicability in everyday scenarios. Consequently, these methods are not suitable to facilitate behavioral changes that could deter heavy drinking or hazardous activities, such as driving.

This study aims to offer a solution by implementing just-in-time adaptive interventions (JITAIs) when users exceed a transdermal alcohol concentration (TAC) of 35 $\mu$g/L on their wrists (equivalent to a blood alcohol concentration (BAC) of 0.05\%), indicating intoxication. Unlike previous research that primarily utilized limited smartphone motion data and conventional machine learning algorithms to classify heavy drinking episodes, our work innovatively integrates multiple sensors—including accelerometer, gyroscope, and heart rate data—from smartwatches over an extended real-world monitoring period. This comprehensive approach enhances the accuracy of intoxication detection and provides a deeper understanding of users' physiological and behavioral states. 
JITAIs aim to provide personalized real-time interventions for users, typically using a mobile device, to mitigate instances of heavy drinking and alleviate associated consequences. Our application also has the potential to address lower levels of alcohol concentration, improving individuals' awareness of their alcohol consumption habits.

In this work, we performed a study to generate a dataset that includes accelerometer, gyroscope, and heart rate data (which has shown a correlation with blood alcohol levels~\cite{10.1042/cs0870225}). Data were collected from 30 participants over a period of three weeks, making it one of the first studies to collect such extensive multi-sensor smartwatch data in a naturalistic setting for intoxication classification. The study was conducted with the consultation of a panel of medical doctors and approved by the Institutional Review Board (IRB) at the University of California Irvine, a R1 university.

\section{Related Work}

Research in methods for assessing alcohol content has evolved significantly. Early methods, such as Nicloux oxidation separation in 1918, required invasive blood sampling~\cite{ANDREASSON19951}. Bodily fluid testing, as gas chromatography, also involves invasive procedures~\cite{gas_chromatography}. The devices of breathing alcohol measure the ratio of alveolar air to ethanol volume~\cite{BrAC_ratios}, while transdermal sensors use electrical or enzymatic principles to measure alcohol through the skin~\cite{transdermal_sensor_survey}. 

Intoxication estimation algorithms differ from traditional methods in that they rely on correlations between intoxication and physiological measurements, established using machine learning~\cite{Paprocki2022}. Techniques using photoplethysmography (PPG) and electrocardiogram (ECG) measurements have shown promising results in intoxication detection~\cite{ecg_ppg_ml}, other approaches involve facial temperature measurements~\cite{facial_temp_ml}, bioimpedances through wearables~\cite{Czaplik2019}, and motion readings from smartphones and smart wearables~\cite{Killian2019LearningTD,bac2}. Integrating these models into non-invasive JITAIs becomes more feasible with advances in sensor technology.


The practicality of using machine learning on smart wearables for JITAIs depends on the computational efficiency of the models. In Human Activity Recognition (HAR), various ML methods have been applied using smartphone data~\cite{Shoaib2014FusionOS} 
, although concerns about power consumption and trade-offs between energy efficiency and model accuracy remain~\cite{Lima2019HumanAR}. Techniques using accelerometer readings and machine learning models such as Random Forest~\cite{bac2}, Convolutional Neural Networks (CNN)~\cite{bac3}, and Support Vector Machines (SVM)~\cite{bac1} have been effective in processing sensor data for intoxication classification. Regression models, including Bayesian Regression Neural Networks (BRNN)~\cite{bac3} and CNN and Bidirectional Long-Short-Term Memory (bi-LSTM) ensembles~\cite{bac4}, have been explored for intoxication regression. However, these methods are computationally expensive in training and inference, limiting their practicality for real-time IoT applications of JITAIs. Recently, there has also been some work that utilizes Hyperdimensional Computing (HDC), also known as Vector Symbolic Architecture (VSA), for the classification of accelerometer intoxication that exceeds the accuracy of previous methods, while allowing efficient single-pass training and online learning in embedded systems for JITAIs~\cite{segura2024enhanced}.

\subsection{Contributions}
In this work, we use state-of-the-art machine learning models to analyze an expanded data set containing accelerometer, gyroscope, heartrate, and TAC data (used as a label for the training process), collected in our own IRB study. We introduce a groundbreaking JITAI mobile application that will run this solution locally on a smartphone/smartwatch pair, while balancing performance with energy efficiency. Our solution addresses the existing research gap in JITAI systems for intoxication by introducing a high-performance model that can conveniently operate on commercial wearables, such as smartwatches.

\section{Data}

We conducted an IRB approved study to investigate new ways to supplant conventional methods of detecting intoxication such as breathalyzers with TAC and BAC sensors. The study was conducted to determine the accuracy of the wearable device in detecting TAC and was carefully designed to be consistent with ethical requirements and regulatory processes. It was subjected to an intensive review by the IRB to ensure the rights and welfare of the participants. Physicians and researchers collaborated to decide the research's scientific value and ethical implications. The IRB reviewed the procedures of the study, the risks, and the benefits before approving.

\subsection{Participant Eligibility}

Particular exclusion and inclusion criteria were applied to select appropriate participants as well as to protect their safety. Participants were required to be between the ages of 21 and 55 years, and needed to possess a valid driver's license and have driven at least 1,000 miles in the past year. Participants were also required to abstain from sedating drugs, such as antihistamines, 24 hours before the first meeting. Excluded were pregnant or breastfeeding persons, individuals with severe medical conditions (such as chronic illness or psychiatric illness), or individuals on drugs with adverse alcohol-altered interactions.

\subsection{Recruitment}
Recruitment strategies included posting notices and online notifications and direct outreach by the research team. Information sessions were held to convey inclusive information about the study purpose, procedures, and participation requirements. All prospective participants signed an informed consent form.

\subsection{Study Procedures and Data Collection}
The study involved four visits with 30 participants over three weeks, two in-person visits and two virtual check-ins. The first visit was an in-person orientation session that lasted about five hours. During that session, participants were given study devices, including a smartwatch, a smartphone, and a TAC sensor bracelet. Measurements for each device are listed in Table~\ref{tab:device_measurements}. The research team walked the participants through how to properly use the devices and how to use them in a manner in which they felt comfortable. Subjects received two 330 ml beers 30 minutes apart, and their BAC was measured several times using a breathalyzer.

\begin{table}[htbp]
\centering
\begin{tabular}{@{}ll@{}}
\toprule
\textbf{Device} & \textbf{Measurements Taken}                      \\ \midrule
Apple Watch Series 8 & Accelerometer data (x, y, z axes) \\
Apple Watch Series 8 & Gyroscope data (x, y, z axes) \\
Apple Watch Series 8 & Heart rate measurements \\
Apple Watch Series 8 & Geolocation \\
BACtrack Skyn Bracelet & TAC measurements every 30 minutes \\
BACtrack C8 Breathalyzer & BAC measurements \\ \bottomrule
\end{tabular}
\caption{Summary of devices and measurements taken during the study}
\label{tab:device_measurements}
\end{table}

In the remaining part of the study, subjects were asked to wear the devices given as part of their day-to-day lives without specific instructions to practice any particular activity: a TAC sensor bracelet, a smartwatch, and a smartphone. The devices recorded data passively, taking physiological readings pertinent to the study goals. 

Throughout the study, participants had weekly virtual check-ins that allowed the research staff to collect user experience feedback regarding the assigned devices. The participants could report any issues they experienced with their assigned devices, ask questions, and receive guidance on troubleshooting technical problems. The regular communication maintained data quality and participant compliance. Following the four-week period, participants had a final physical visit to return the borrowed research devices and received financial compensation for participation.


\subsection{Data Preprocessing}

The resulting data set included gyroscope, accelerometer, heart rate data, and the corresponding TAC values for the 30 participants. Data were recorded at a sampling frequency of 50 Hz, with separate x, y, and z components for accelerometer and gyroscope readings. The data set also included heart rate values, TAC measurements (in $\mu$g/L), and a binary indicator that indicates whether the TAC exceeded 35 $\mu$g/L, which indicated a moderate drinking session~\cite{bactrack_skyn_tac}.

To organize the data effectively, we divided the data for each participant into user "sessions," defined as periods of continuous recording of more than one minute. Each session includes all sensor data collected when all three devices, TAC sensor, smartwatch, and smartphone, were operational simultaneously. Gaps in recording, such as instances where one device was not collecting data, were excluded from the data set. Figure~\ref{fig:session} illustrates an example of TAC sensor data within a session.

\begin{figure*}[htbp!]
\centering
\includegraphics[width=\textwidth]{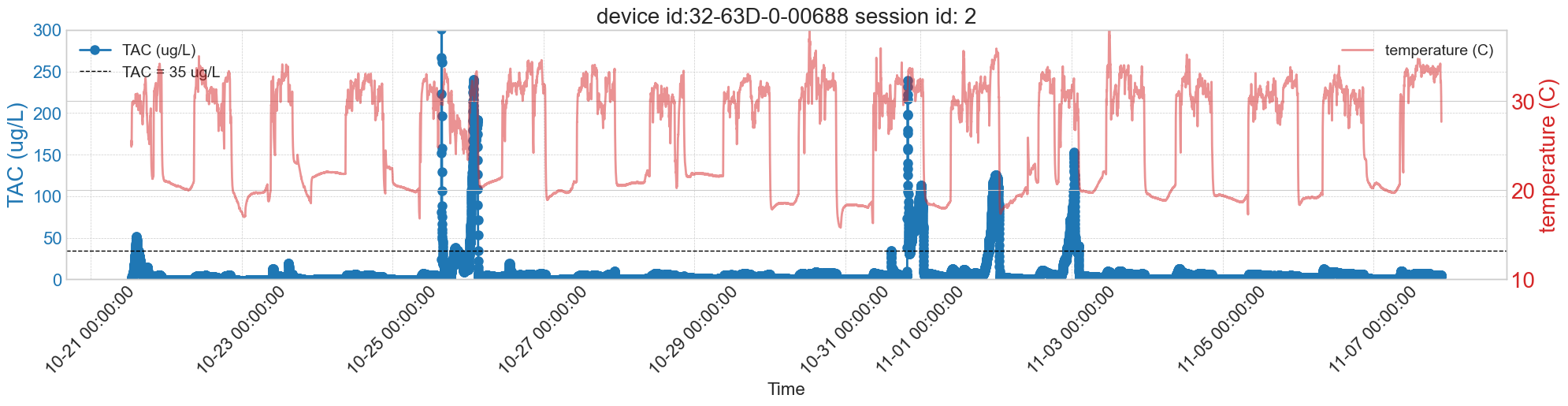}
 \caption{Example user session with TAC spikes above 35$\mu$g/L}
 \label{fig:session} 
\end{figure*}

We focused on the participants who presented drinking sessions with TAC readings exceeding 35 $\mu$g/L, narrowing the focus to 14 individuals out of the initial 30 participants. This was due to the high tolerance of some of the recruited participants and our desire to not lower the TAC threshold for classification. The drinking sessions of each participant were further analyzed in 20-second windows, focusing on both the time and frequency domains to identify key features of the accelerometer and gyroscope signals. Analysis revealed that most frequency peaks occurred below 5 Hz. Based on this observation, we applied a low-pass filter and chose a reduced sampling frequency of 40 Hz to preserve the critical components of the signal while reducing computational complexity.

\section{Methods}
Our primary task was to determine whether a subject was under the influence of alcohol or not by only using wearable data~\cite{rosa2020wearable}. To do this, we used the data that we have gathered to evaluate seven models to assess their accuracy and applicability to our problem. These models have been selected based on state-of-the-art learning, state-of-the-art time series classification~\cite{bagnall2017great}, state-of-the-art blood alcohol classification, and usability on embedded devices (energy, memory, and time efficiency)~\cite{verhelst2019embedded}.

The first model was an SVM (using the radial basis function kernel), which has shown high accuracy results while being computationally fast and efficient~\cite{cortes1995support, zhang2021svm}. In our implementation, we included a feature extraction step using a neural network comprising a flattening layer, a dropout rate of 0.1, and a linear layer reducing the input features to 128 dimensions followed by a ReLU activation. This was followed by a linear SVM classifier acting as the output layer. The second model was a LightGBM, another model that has shown excellent accuracy results in classification while being computationally efficient~\cite{ke2017lightgbm, li2022lightgbm}. We configured the LightGBM with a learning rate of 0.1, using 32 leaves per tree, a maximum depth of 4, and a total of 5 estimators. The regularization parameters $\alpha$ and $\lambda$ were set to 0.5 and the minimum number of data in a leaf was set to 4. The third model was a bi-LSTM, which could capture the context of past states, which was very useful for our time-series classification~\cite{schuster1997bidirectional, wang2020bilstm}. Our bi-LSTM model consisted of four layers with 128 hidden units each, using a bidirectional architecture and a dropout rate of 0.1. We also implemented an attention mechanism to focus on relevant time steps, and the model's output was passed through fully connected layers to produce the final prediction. The fourth model was the Gated Recurrent Unit (GRU), which could also capture information from past events~\cite{cho2014gru}, but was more efficient than an bi-LSTM~\cite{kim2023gru}. Our GRU model had a single layer with 64 hidden units and a dropout rate of 0.1. Similar to the bi-LSTM, we incorporated an attention mechanism. The final output was processed through fully connected layers leading to a sigmoid activation for classification. Our fifth model was the Transformer, which could use a self-attention mechanism to capture long-range dependencies, which was useful for our time-series classification~\cite{vaswani2017transformer, liu2024transformer}, but could require more computational resources.  We implemented a Transformer encoder with 2 layers and an embedding dimension of 128, incorporating positional encoding and dropout to prevent overfitting. Our sixth model was an One-dimensional CNN (1D-CNN), which used convolution operations on a single dimension, which was well suited for time series~\cite{lecun1998gradient, kiranyaz20211dcnn} and could be more efficient than recurrent networks. The 1D-CNN model included three convolutional layers with 32 filters each, using kernel sizes of 3, 5, and 7 respectively. Our seventh and final model utilized HDC, which has shown great results in time-series applications and detecting blood alcohol content while also being very energy efficient ~\cite{segura2024enhanced}. We used the key-value encoding and RefineHD algorithms described in \cite{segura2024enhanced} to develop a HDC model for this experiment, with hypervectors of dimensionality 3000.

\section{Mobile and Smart-watch Application}
We developed a mobile-watch application for our IRB study protocol to help users track their daily alcohol consumption. As illustrated in Figure~\ref{fig:currmobileapp}, the application allowed users to select and log the specific drinks they consume. Additionally, users could record detailed information about their drinking episodes, including the quantity and type of beverages consumed. The application also collected continuous data from the accelerometer, gyroscope, and heart rate sensors. Combined with TAC measurements obtained from the BACtrack Skyn bracelet, these data streams enabled us to compile a comprehensive dataset for our study.


The results presented in this study allowed the design of a mobile app that provides a diary of the drinks taken during the day/month/year and can identify when the user has surpassed the legal limit of intoxication without any user input. This could help users better track how alcohol affects them and better limit and control their alcohol consumption. A screenshot of the statistics page of our current application is shown in Figure~\ref{fig:mobileapp}, where the features mentioned above are shown. 


\begin{figure}[htbp]
\centering
\begin{subfigure}[t]{0.48\columnwidth}
\centering
\includegraphics[width=\linewidth]{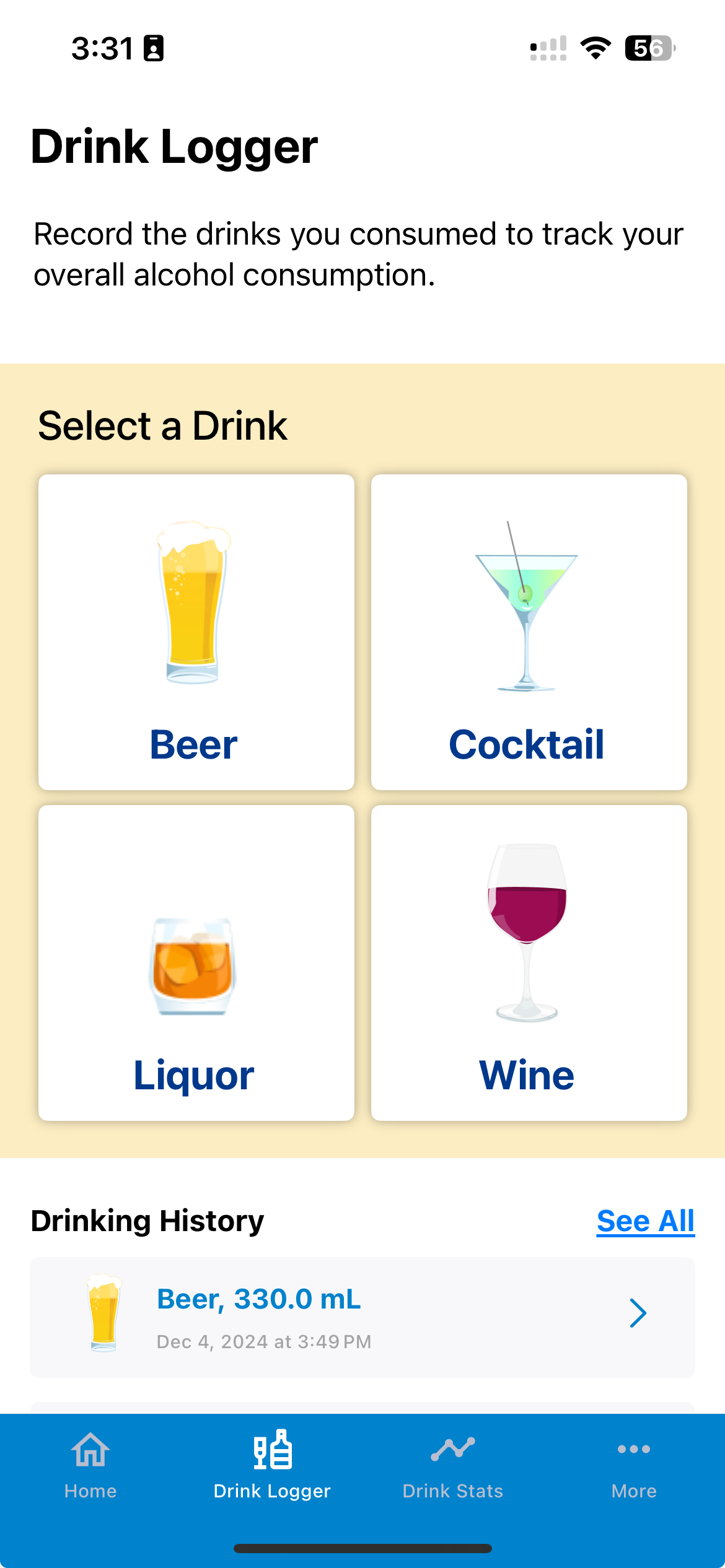}
\caption{\label{fig:currmobileapp} Drink logger.}
\end{subfigure}
\hfill
\begin{subfigure}[t]{0.48\columnwidth}
\centering
\includegraphics[width=\linewidth]{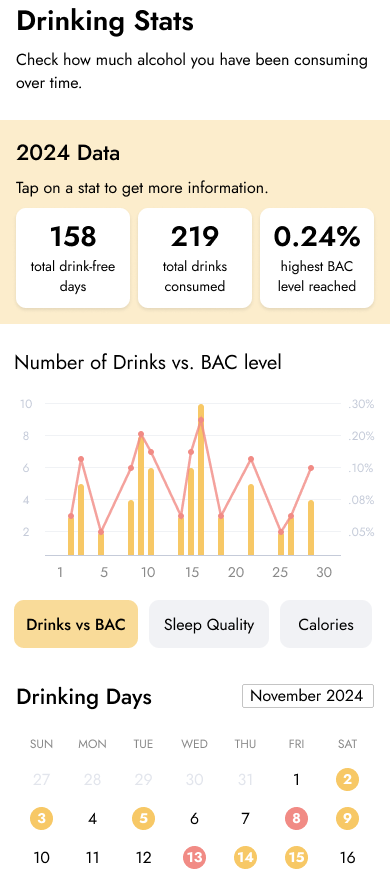}
\caption{\label{fig:mobileapp} Statistic application page.}
\end{subfigure}
\caption{\label{fig:mobileApp} Mobile application screens.}
\end{figure}

Moreover, our application could also send JITAI notifications when the user reaches certain levels of BAC. In Figure~\ref{fig:intoxicated}, we show the just-in-time intervention when the user surpasses the legal limit of BAC; in this case, our app would show a button to open a ride-share app to help the user make the right decision of not driving while being under the influence.

\begin{figure}[htbp]
  \includegraphics[width=.35\columnwidth]
    {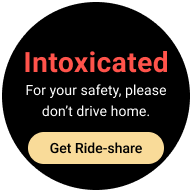}
    \hfill
\includegraphics[width=.35\columnwidth]
    {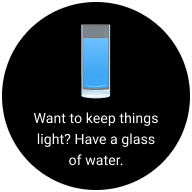}

  \caption{\label{fig:intoxicated} Smartwatch JITAI intervention.}
\end{figure}

\subsection{Outline of Experiments}
The data diversity of our data set was a significant issue in our experiments, so an approach to generalize our models as much as possible was necessary. To maximize the generalization of our models with the 14 subjects we had chosen, we group users by k-means into 3 categories according to their overall TAC values: high, medium and low, as shown in Figure~\ref{fig:tac_clustering}. One subject from each category was selected to form the test set. These users were completely held out during training to have an unbiased estimate of model performance on new data from each category.

\begin{figure}[htbp!]
\centering
\includegraphics[width=\columnwidth]{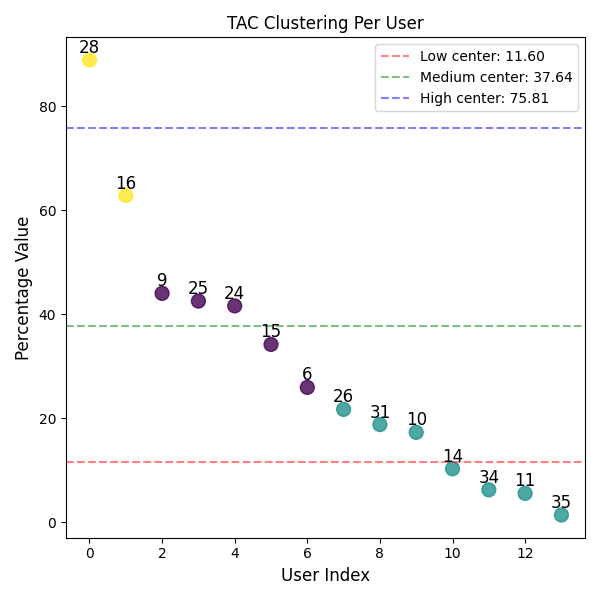}
 \caption{Clustering of Users based on TAC value}
 \label{fig:tac_clustering} 
\end{figure}

We used a stratified group k-fold cross-validation strategy for the remaining 11 subjects with three folds. The particular assignment of subjects to training and validation sets for each fold is detailed below.
\begin{itemize}
    \item Fold 1:
    \begin{itemize}
        \item Training set user IDs: 34, 35, 6, 10, 11, 15, 28, 31; 
        \item Validation set user IDs: 24, 9, 14. 
    \end{itemize}
    \item Fold 2:
    \begin{itemize}
        \item Training set user IDs: 35, 6, 9, 10, 14, 24, 28
        \item Validation set user IDs: 34, 11, 31, 15.
    \end{itemize}
    \item Fold 3:
    \begin{itemize}
        \item Training set user IDs: 34, 9, 11, 14, 15, 24, 31;
        \item Validation set user IDs: 10, 35, 28, 6.
    \end{itemize}
    \item Test set user IDs: 16, 25, 26.
\end{itemize}

The cross-validation strategy enabled both model training and validation in various sets of participants with user uniqueness between different sets. 
We were targeting the model's generalizability by ensuring that every fold contained a proportionate subset of TAC levels and that no users were present in both the training and validation sets within a given fold. The final model was tested using the test set. This particular set enabled a complete estimation by testing the model's performance on completely new subjects, thereby assessing its capacity to generalize from the training and validation datasets used. The application of this partitioning technique was essential in creating a sound model that is proficient at correctly predicting fluctuations in the TAC level in different individuals with different drinking habits. It further enabled the testing of the model's performance in real-life conditions, where it has to predict outcomes for new subjects based on information acquired from other subjects.

We executed the trained models saved from our cross-validation experiment on an edge device to demonstrate the feasibility of performing inference. We utilize the Pytorch Executorch library~\cite{executorch2023} to run models using a prototype Android application on a Samsung Galaxy S20 with a Qualcomm Snapdragon 865 processor.

\section{Results and Discussion}
In our study, we evaluated the performance of state-of-the-art algorithms for time-series classification. Our results are shown in Table~\ref{tab:metrics}. Among the models compared, LightGBM and HDC performed the best according to the ROC-AUC metric with scores of 0.746 and 0.740, respectively. However, LightGBM classified all sober examples as drunk, making the model unusable. 

\FloatBarrier
\begin{table*}[htbp]
    \vspace*{0.1in} 
    \centering
    \begin{tabular}{@{}lccccccc@{}}
        \toprule
        \textbf{Model} & \textbf{ROC-AUC} & \textbf{PR-AUC} & \textbf{Accuracy} & \textbf{Sober Accuracy} & \textbf{Drunk Accuracy} & \textbf{F1 Score} & \textbf{Threshold}\\
        \midrule
        1D-CNN       & \textbf{0.748480} & 0.726916 & \textbf{0.760990} & 0.846154 & 0.615595 & 0.655499 & 0.260000\\
        bi-LSTM      & 0.717478 & 0.689900 & 0.369378 & 0.000000 & \textbf{1.000000} & 0.539483 & 0.020000\\
        GRU          & 0.684015 & 0.690456 & 0.425720 & 0.159054 & 0.880985 & 0.531244 & 0.190000\\
        HDC          & 0.744281 & \textbf{0.734389} & 0.760738 & 0.828926 & 0.644323 & \textbf{0.665489} & \textbf{N/A}\\
        LightGBM     & 0.746116 & 0.702795 & 0.369378 & 0.000000 & \textbf{1.000000} & 0.539483 & 0.001000\\
        SVM          & 0.255836 & 0.263962 & 0.630622 & \textbf{1.000000} & 0.000000 & 0.000000 & \textbf{0.500000}\\
        Transformer  & 0.529188 & 0.422930 & 0.369378 & 0.000000 & \textbf{1.000000} & 0.539483 & 0.005036\\
        \bottomrule
    \end{tabular}
    \caption{Performance metrics per model type}
    \label{tab:metrics}
\end{table*}

The HDC and 1D-CNN models presented the most balanced performances in the updated analysis. The HDC model achieved an overall accuracy of 76.1\%, sober accuracy of 82.9\%, and drunk accuracy of 64.4\%. It also achieved the highest F1 score of 0.665, indicating a higher balance between recall and precision. Similarly impressive, the 1D-CNN model attained the highest ROC-AUC of 0.748 and the best overall accuracy of 76.1\%, with a strong balance between sober accuracy (84.6\%) and drunk accuracy (61.6\%), along with a competitive F1 score of 0.655. The bi-LSTM model showed an ROC-AUC of 0.717, but its performance was unbalanced with 0\% sober accuracy and 100\% drunk accuracy. Since there is a requirement for the models to correctly classify both drunk and sober cases, both the HDC and 1D-CNN models emerge as the most successful and balanced of the models under consideration.

In terms of model sizes and computational complexity, the HDC model was the largest at 36.7 megabytes (MB), which may affect deployment in very resource-limited settings, but is more than small enough for smartphones or smartwatches. The 1D-CNN model was extremely lightweight, having a model size of only 0.035 MB. This combination of high performance and small size makes the 1D-CNN particularly attractive for applications where computational resources are very constrained. The GRU model offered a trade-off, with a small model size of 0.082 MB and a good drunk accuracy of 88.1\%, though with a lower overall accuracy of 42.6\%. Trading off computational efficiency and performance metrics, there exists a model size accuracy trade-off.

From the metrics, the two most balanced performing models were HDC and 1D-CNN. Although larger in size, the HDC model offers excellent balanced performance, while the 1D-CNN provides the best combination of high accuracy and computational efficiency. In very resource-constrained applications, the 1D-CNN presents an ideal solution with virtually no compromise on performance despite its significantly smaller size and complexity.

\subsection{Benchmarking on Android with Executorch}

We tested some of the PyTorch-based models on a Samsung Galaxy S20 using Executorch for 100 iterations. The Samsung Galaxy S20 had a Qualcomm SM8250 Snapdragon processor and 8 gigabytes (GB) of Random Access Memory (RAM). Our results are shown in Table~\ref{tab:android_metrics}. The 1D-CNN model was very efficient with a mean inference time of 0.0121 seconds, small memory usage (52.3 MB), and 0.315 Watts (W) on average, which was able to process data within the desired time interval without any issues. The SVM model was even quicker, with an inference time of 0.0034 seconds, 83.5 MB of memory usage and 0.207 W on average. The HDC model consumed 0.0852 seconds per inference, 328 MB of memory (4\% of total memory), and 0.481 W on average, and was well within the 20-second window. The Transformer model was slower at 0.3285 seconds, 90.1 MB of memory, and 0.599 W on average, which was acceptable considering the huge delay between data arrival. 

We were unable to test the LightGBM model as it was not implemented in PyTorch and, thus, was not compatible with Executorch. Furthermore, attempts to export the bi-LSTM and GRU models to Executorch were unsuccessful because Executorch was unable to unroll their graphs, which led to unsuccessful exports. These limitations indicate that both model performance and compatibility should be considered when deploying models on mobile devices.

\begin{table}[htbp]
\centering
\begin{tabular}{@{}llll@{}}
\toprule
\textbf{Model} & \textbf{Avg Time (s)} & \textbf{Avg Mem (MB)} & \textbf{Avg Power (W)} \\ \midrule
1D-CNN & 0.0121 & 52.3 & 0.315 \\
HDC & 0.0842 & 328 & 0.481 \\
SVM & 0.0034 & 83.5 & 0.207 \\
Transformer & 0.3285 & 90.1 & 0.599 \\ \bottomrule
\end{tabular}
\caption{Summary of hardware inference metrics on Samsung Galaxy S20, 100 iterations}
\label{tab:android_metrics}
\end{table}

\section{Future Work}

Future work will focus on expanding the dataset to include a broader range of users, enabling the development of a more robust and generalized model that can potentially consider users' profiles and demographic variables. Additionally, implementing the application with the 1D-CNN, or HDC model using C and vectorized operations, can improve efficiency and reduce energy consumption, increasing its feasibility on resource-constrained devices.

The final application is envisioned as an affordable and accessible tool for the general public. It may allow users to monitor instances of intoxication and provide just-in-time interventions by notifying them of potential intoxication events. These notifications are intended to support users in making informed decisions during critical moments, ultimately promoting healthier drinking habits.

Integrating machine learning with JITAIs presents a promising avenue to advance real-time personalization and intervention adaptability. Future research could explore the scalability of this approach across diverse populations and environments and its application to other behavioral health challenges. Addressing key issues such as data privacy and conducting rigorous validation in real-world scenarios will be essential to achieve widespread adoption. This work aims to promote healthier lifestyles and enhance overall well-being by advancing technology-driven solutions to public health challenges.

\section{Conclusions}

Excess alcohol consumption poses significant risks to individual health and well-being. JITAIs have emerged as a highly promising solution among various strategies to encourage responsible drinking. This study tested multiple state-of-the-art algorithms to evaluate the feasibility of IoT-enabled JITAIs to address challenges related to alcohol consumption. The results demonstrate that the HDC model used achieved the most balanced performance compared to the other models evaluated.

These results recognize the prospects of 1D-CNN and HDC models for actual deployment in IoT-enabled JITAIs to encourage moderate drinking habits. Both classifiers, in addition to realizing balanced and precise performance, are also efficient enough to be executed on resource-limited devices such as smartphones, making it viable for ongoing real-time monitoring. This opens up opportunities for the delivery of personalized interventions capable of successfully alerting the individual during times of heightened risk, thereby reducing the harmful effects of heavy drinking. Through the use of cutting-edge machine learning algorithms and wearable sensors, the system guarantees an innovative proactive approach in the public health effort towards preventing alcohol harm.



\bibliographystyle{IEEEtran}
\bibliography{bibliography}

\end{document}